\documentclass{article}
\usepackage{arxiv}

\usepackage{tabularx}
\usepackage{multirow}
\usepackage{array}
\usepackage[ruled,vlined]{algorithm2e}

\usepackage[dvipsnames]{xcolor}
\usepackage{censor}
\usepackage[utf8]{inputenc} 
\usepackage[T1]{fontenc}    
\usepackage{hyperref} 
\usepackage{scalerel,stackengine}
\usepackage{ragged2e}
\usepackage{xurl}
\stackMath
\newcommand\reallywidehat[1]{%
\savestack{\tmpbox}{\stretchto{%
  \scaleto{%
    \scalerel*[\widthof{\ensuremath{#1}}]{\kern-.6pt\bigwedge\kern-.6pt}%
    {\rule[-\textheight/2]{1ex}{\textheight}}
  }{\textheight}%
}{0.5ex}}%
\stackon[1pt]{#1}{\tmpbox}%
}
\definecolor{note}{HTML}{235c1f}
\usepackage{graphicx}
\usepackage[export]{adjustbox}
\newcommand\Tstrut{\rule{0pt}{2.1ex}}       
\newcommand\Bstrut{\rule[-0.9ex]{0pt}{0pt}} 
\newcommand{\TBstrut}{\Tstrut\Bstrut} %
\newcommand{\foo}{\hspace{-2.3pt}\textcolor{blue}{$\bullet$} \hspace{5pt}}
\usepackage{colortbl}
\usepackage{paralist}
\usepackage{bbm}
\usepackage{url}            
\usepackage{booktabs}       
\usepackage{amsfonts}       
\usepackage{nicefrac}       
\usepackage{microtype}      
\usepackage{lipsum}	
\usepackage{pifont}

\usepackage{subcaption}
\usepackage{tikz}
\usetikzlibrary{shapes.geometric}

\usepackage{relsize}
\usepackage{mathtools}

\usepackage{multirow}
\usepackage{multicol}
\usepackage{paralist}
\usepackage{float}
\usepackage[numbers]{natbib}

\title{Gender Representation and Bias in Indian Civil Service Mock Interviews}
\author{
Somonnoy Banerjee\thanks{Somonnoy Banerjee and Sujan Dutta are equal contribution first authors. Ashiqur R. KhudaBukhsh is the corresponding author.} \\
  \small{Rochester Institute of Technology}\\
  \texttt{sb7238@rit.edu} \\
  \And
  Sujan Dutta$^*$\\
  \small{Rochester Institute of Technology}\\
  \texttt{sd2516@rit.edu} \\
\And
Soumyajit Datta \\
  \small{Rochester Institute of Technology}\\
  \texttt{sd3528@rit.edu} \\
 \And
Ashiqur R. KhudaBukhsh \\
  \small{Rochester Institute of Technology}\\
  \texttt{axkvse@rit.edu} \\
}

\begin{document}
\maketitle

\begin{abstract}
This paper makes three key contributions. First, via a substantial corpus of 51,278  interview questions sourced from 888 YouTube videos of mock interviews of Indian civil service candidates, we demonstrate stark gender bias in the broad nature of questions asked to male and female candidates.  Second, our experiments with large language models show a strong presence of gender bias in explanations provided by the LLMs on the gender inference task. Finally, we present a novel dataset of 51,278  interview questions  that can inform future social science studies.  

\end{abstract}

\section{Introduction}


\renewcommand{\labelitemi}{\textcolor{black}{\textbullet}}
\begin{compactitem}
    \item  \textcolor{red!}{\textit{What is the age of your kids?}}
    \item \textcolor{red}{\textit{Provide tips to keep your kids busy.}}
    \item \textcolor{red}{\textit{Who is there to handle the kids in your absence?}}
\end{compactitem}


\begin{compactitem}
    \item \textcolor{blue}{\textit{How is the poverty line defined now?}}
    \item \textcolor{blue}{\textit{What is the role of Sanchi Stupa in the national emblem of India?}}
    \item \textcolor{blue}{\textit{What was the basic philosophy of Kautilya in Political Science?}}
\end{compactitem}

The surprising thread connecting these two contrasting sets of questions is that they appeared in the preparatory UPSC mock interviews organized by the same coaching institute. However, with one key difference — the first set was asked of a female candidate, and the second one was asked of a male candidate. 

Tracing back its origin to the Imperial Civil Service (ICS)~\cite{cornell2020imperial}, Indian Administrative Service (IAS) has a long and decorated history that shaped the India we see today. The IAS holds significant influence in Indian governance, forming the administrative backbone of the world's largest democracy. Due to its strong influence on public policy, the Civil Services Examination, organized by the Union Public Services Commission (UPSC), is one of the most competitive exams in India, with around a million aspirants applying every year.  The exam consists of multiple written tests with the final phase involving an interview/personality test.  



A growing market of coaching institutes has emerged providing coaching to these millions of aspirants. Many of these institutes have a strong online presence and have published mock interview videos of the personality test of the top candidates on their YouTube channels for broader accessibility of their training materials. Online educational resources often serve as a great leveler for broadening participation~\cite{chtouki2012impact,hansen2015democratizing}. However, unlike traditional educational resources~\cite{lucy2020content,parashar2020evaluating}, little or no computational audits for bias exists for such resources.


While gender bias has a rich and extensive literature in diverse social and computational settings that include hiring decisions~\cite{marlowe1996gender}, machine translation~\cite{ghosh2023chatgpt}, movie transcripts and narrative tropes~\cite{gala-etal-2020-analyzing}, interview processes~\cite{kane1993interviewer}, word embeddings~\cite{DBLP:journals/pnas/GargSJZ18}, academic textbooks~\cite{blumberg2008invisible}, and political interruptions~\cite{DBLP:conf/ijcai/YooWLKK22}, UPSC mock interviews present a rare lens to the interview process of one of the most coveted job positions in India and to the best of our knowledge, no comprehensive AI-powered analyses have scrutinized gender bias in these interviews. \textit{Is it possible that beneath the veneer of seemingly innocuous assortment of interview questions on public policy, international relations, cutting edge technologies, and social studies, lies a biased pattern where women are consistently asked different questions than their male counterparts?} Via a substantial corpus of 888 mock civil service and administrative service mock interview videos published by 14 well-known coaching institutes, this paper seeks to conduct a thorough investigation of this research question.

Our mixed-method analyses reveal that (1) women are almost thrice as likely as men to be asked questions about gender equality or family; (2) while the candidates in mock interviews show reasonable gender distribution (65.43\% male and 34.57\% female), the interview panels exhibit significantly more skewed gender distribution; finally, (3) large language models exhibit societal biases in their explanations when tasked with gender inference from interview transcripts.




To summarize, our contributions are the following:


\renewcommand{\labelitemi}{\textcolor{blue}{\textbullet} }

\begin{itemize}
    \item \textbf{\textit{Resource}}: We compile a substantial corpus of 51,278  interview questions sourced from 888 UPSC mock interviews conducted by 14 prominent coaching institutes. These questions, along with the video transcripts, will enable social science researchers to investigate other important questions relating to the topics featured in these interviews. 
    \item \textbf{\textit{Social}}: Our analyses reveal that women face substantially more questions around gender equality, family, and women empowerment and considerably fewer questions on international affairs, world politics, and sports suggesting a strong presence of gender stereotypes. 
    \item \textbf{\textit{Methodological}}: In an experiment to infer gender from interview transcripts, we observe that several cutting-edge LLMs exhibit stereotypes in their explanations that point to deep-seated gender bias in cutting-edge LLMs.  
\end{itemize}

\section{Dataset}

\subsection{Step 1: Identifying Relevant Videos} We first construct a set of relevant videos depicting mock interviews conducted with candidates preparing for Civil Services examinations. We consult 14 YouTube channels managed by prominent training institutes (see \textbf{SI} for further details). These channels have a strong viewer engagement with 2,378,857 $\pm$ 4,259,079 subscribers and median video views of 42 million. We use publicly available YouTube API and collect all videos from these channels. We next filter in all videos whose titles contain the phrase \texttt{mock interview}. To avoid YouTube shorts and promotional videos, we discard any video that lasts less than 10 minutes. This yields a set of 888 videos, denoted by $\mathcal{V}$.  We note that $\mathcal{V}$ not only includes mock interviews of candidates preparing for the Civil Services positions of the Union/Central government but also state governments like Uttar Pradesh, Bihar, and Rajasthan.



When we contrast the academic background distribution of a random sample of 200 candidates from $\mathcal{V}$ (obtained through manual inspection of videos) with ground truth sourced from official UPSC statistics. we observe that $\mathcal{V}$ is a representative sample of successful UPSC candidates and is consistent with the academic distribution background of the recommended candidates (\textbf{SI} contains the Table).



\subsection{Step 2: Obtaining Interview Transcripts}
663 videos (74.66\%) in $\mathcal{V}$ have creative commons license.  
For these videos, we generate transcripts from the audio information using Whisper OpenAI. For the remaining videos, we first obtain the transcript using publicly available YouTube API. YouTube official transcripts do not have punctuation such as question mark. We use \texttt{GPT-3.5} to add appropriate punctuation to the transcript. The transcribed corpus, $\mathcal{D}$, consists of 4.5 million tokens. $\mathcal{D}$ consists predominantly of conversations in English. There are, however, a few instances where the interview had conversations in both English and Hindi. We observe that when the conversations switched to Hindi, the ASR system often repeats its previous generations. To account for this, we remove sentences that repeat three or more times in a row. A manual inspection on a small subset of videos confirms that the transcripts have high fidelity with actual audio even including accurate transcription of Indian names if mentioned in the audio. Our use of these publicly available interviews of public officials  hosted on public social web platforms for research purpose comes under the purview of fair use.  

\subsection{Step 3: Gender Inference of Interview Candidates}

Any contrastive study involving gender requires partitioning instances based on gender information. 
However, annotating image or videos for race and gender information is often treated as insignificant, irrefutable, and apolitical process~\cite{scheuerman2020we}.  We adopt a sociotechnical approach for gender inference of the interview candidates considering multiple sources. We obtain consensus labels from two human annotators who had access to the (1) video titles (titles list candidate names); (2) video transcripts; (3) video thumbnails; and (4) videos. Indian personal names often indicate gender~\cite{sharma2005panorama}, and computational studies aimed at gender inference from Indian personal names exists~\cite{gulati2015extracting}. The annotation process was informed by subcultural naming conventions in last names as well (for instance, \texttt{Kaur}, meaning princess, is a Punjabi last name only for females~\cite{kaur2019gap}). The annotators considered the video frames, videos, and audio transcript and share that formal male (suit) and female attire (97.06\% of the female candidates wore sari or kurti); domain-specific knowledge (e.g., if a candidate received gender-isolated education); and of course the pronouns with which the candidate is being referred to -- contributed to this annotation process. Overall, we identify 581 videos ($\mathcal{V}_\textit{male}$) of male candidates and 307 videos ($\mathcal{V}_\textit{female}$) of female candidates. These set of labels is denoted by $\mathcal{L}_\textit{comprehensive}$. 

Barring recent candidates who are still receiving administrative training, most of these interview candidates have already joined as highly visible public officials. We conduct online search on the candidate names and identify news articles, interview videos (as a celebrated exam topper) and tally our initial annotation with gendered pronouns used in these articles. This process also uncovered further corroborating evidence (e.g., one candidate was a beauty pageant winner and a female model). Finally, the resumes of the candidates already in Indian Administrative Service are publicly available. The gender information is listed in these resumes. We consider this information to be the closest to self-determined gender which we consider the ultimate ground truth that we do not possess. Our initial gender inference from videos tally 100\% with gender inference conducted through this process. 

We also conduct gender inference using large language models and observe interesting stereotypes and biases in their explanations which we discuss in the results section. 

\subsubsection{Gender Inference from Names Only.}
Separate from the two annotators who constructed $\mathcal{L}_\textit{comprehensive}$, we task another annotator who is an expert social scientist with inferring gender solely from the candidate names. The Cohen's $\kappa$ with $L_\textit{comprehensive}$ is 0.81. The human annotator struggled with gender-neutral names. On a similar task, we observe \texttt{GPT-3.5} \footnote{https://openai.com} and \texttt{Claude-3.5-Sonnet} \footnote{ https://www.anthropic.com/} achieve superior Cohen's $\kappa$ of 0.91 and 0.89, respectively, establishing that (1) multiple sources (e.g., image, news articles, resumes) contribute to more robust gender inference; and (2) these LLMs have cultural grounding of Indian names. 




\subsection{Step 4: Sets of Interview Questions}

From $\mathcal{D}$, we construct $\mathcal{Q}$ consisting of sentences that end with a question mark as the set of questions asked of the candidates. To preserve the context of the questions, we also included the sentence that appeared before each question. We acknowledge that this is a high-recall approach with certain caveats. For instance, this set will include clarification questions asked by the candidates and exclude imperative sentences (e.g., \textit{please give a brief introduction}). A manual inspection of randomly sampled 100 questions reveals that 3 are clarifying questions asked by the candidates.  $\mathcal{Q}_\textit{male}$ and $\mathcal{Q}_\textit{female}$ denote all the questions asked on male and female candidates, respectively.

\section{Related Work}



Gender bias has a rich and extensive literature in diverse social and computational settings that include hiring decisions~\cite{marlowe1996gender}, machine translation~\cite{ghosh2023chatgpt}, movie transcripts~\cite{DBLP:journals/patterns/KhadilkarKM22b}, interview processes~\cite{kane1993interviewer}, word embeddings~\cite{DBLP:journals/pnas/GargSJZ18}, academic textbooks~\cite{blumberg2008invisible}, and political interruptions~\cite{DBLP:conf/ijcai/YooWLKK22}. While gender bias has a rich literature, barring a few instances~\cite{madaan2018analyze,DBLP:journals/patterns/KhadilkarKM22b,IJCAI2023Divorce}, AI-powered, computational analyses of gender and societal biases in the Indian context are rather underexplored. Our work contrasts with existing lines of work (1) in terms of domain (Civil Service interview versus gender inequality in Bollywood~\cite{madaan2018analyze,DBLP:journals/patterns/KhadilkarKM22b} and divorce court proceedings~\cite{IJCAI2023Divorce}); and (2) nuanced analyses of bias in LLM explanations.  


Gender bias in hiring and the interview process is a significant barrier to achieving workplace diversity and inclusion. Davison and Burke \cite{davison2000sex} conducted research spanning from the 1970s to the present and showed consistent highlights of gender discrimination in hiring as a persistent issue in today's workplace. Despite a growing belief in competence equality over time, as noted in a cross-temporal meta-analysis by Eagly et al. \cite{eagly2020gender}, recent research by Lippens et al. \cite{lippens2023state} reveals the complexity of gender discrimination in hiring, with both men and women experiencing discrimination in certain contexts. The advantages of male and female candidates are influenced by various demand-side factors, including job characteristics traditionally associated with specific genders and how well candidates conform to gender norms. Castaño et al. \cite{castano2019can} found that women who take on roles traditionally seen as masculine are viewed as cold and driven, while those who align with feminine roles are seen as less capable. Men don't usually face this type of bias. As a result, even when women perform as well as their male counterparts, they are often rewarded less in prestigious jobs. This contributes to a steady drop in female representation at higher levels of corporate and organizational leadership \cite{joshi2015can}. 


A range of research studies has explored gender bias in explainability, highlighting that bias in AI systems can manifest in the explanations provided by these models, affecting their transparency and fairness. For instance, Huber \textit{et al.} \cite{huber2023explainability} explore potential gender bias in explainability tools used in face recognition systems. These tools, designed to provide insights into ML models, might exhibit gender-based bias, leading to signs of biased decisions in critical applications like face recognition. Shrestha and Das \cite{shrestha2022exploring} conduct a systematic review to identify gender biases in ML and AI academic research.

\begin{figure}[t]
   \centering
   \includegraphics[width=0.7\columnwidth]{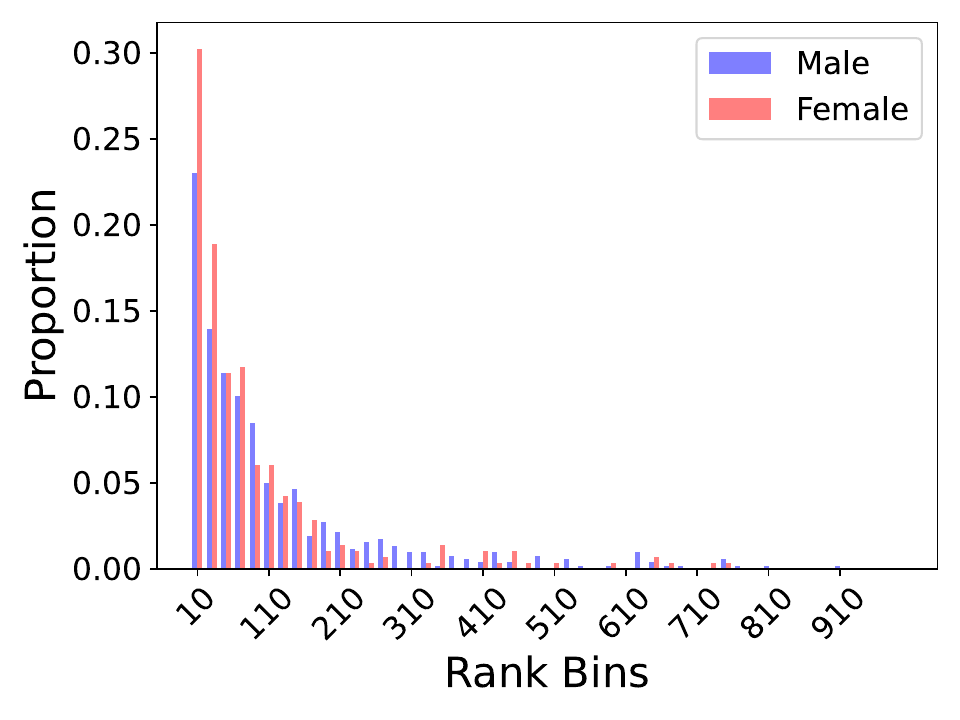}
   \caption{Distribution of records based on rank and gender. Rank information is obtained from the video title.}
   \label{fig:ranking}
\end{figure}

\section{Results and Discussion}


We start with an important point for the readers as they learn about our findings regarding gender representation and bias: \textbf{\textit{the female candidate pool in the mock interviews is as strong as (if not slightly better than) their male counterparts.}} As already mentioned, of the multiple phases in the UPSC exam, the final phase is the personality test. Figure~\ref{fig:ranking} summarizes the candidates' overall performance taking into account the written as well as the personality test. We further note that no significant differences exist in the average number of questions and interview duration between male and female candidates (\textbf{SI} contains details).  

\subsection{Representation}


\noindent\textbf{\textit{Observation 1: Gender representation in YouTube mock interviews is not far from real-world representation.}} As already noted, $\mathcal{V}_\textit{male}$ represents 65.43\% of our candidate pool while $\mathcal{V}_\textit{female}$ represents the remaining 34.57\%. Hence, the gender representation of $\mathcal{V}$ is visibly skewed. However, the imbalance is not far from real-world gender imbalance in UPSC recommendations. Real-world data indicates that the percentage of women candidates recommended by the UPSC has increased from 24\% in 2018 to 34\% in 2022\footnote{\url{https://indianexpress.com/article/jobs/civil-services-ias-upsc-jobs/women-top-upsc-civil-services-higher-performance-8625808/}}. 


\noindent\textbf{\textit{Observation 2: The interview panels exhibit stark gender imbalance.}} We observe that the candidates refer to male panelists and the female panelists with the formal honorific \texttt{sir} and \texttt{ma'am} (short form of \texttt{madam}), respectively. Let $\mathcal{N}_\textit{s}$ and $\mathcal{N}_\textit{m}$ denote the count of the usage of \texttt{sir} and \texttt{ma'am}, respectively. We compute the male honorific ratio (MHR) $\frac{\mathcal{N}_\textit{s}}{\mathcal{N}_\textit{s} + \mathcal{N}_\textit{m}}$. A value closer to 1 indicates a predominantly male panel whereas a value closer to 0.5 indicates a gender-balanced panel. We observe a value of 0.81 for \texttt{MHR} indicating a predominantly male panel composition. A manual inspection of randomly sampled 200 videos aligns with this observation.   


\subsection{Bias in Discourse and Questions}

Our findings from a series of experiments with varying sophistications indicate considerable gender bias.

\subsubsection{Unigram Differential Analysis.}
\label{sec:nf}
A unigram differential analysis illustrates the difference between the discourse in $\mathcal{D}_\textit{male}$ and $\mathcal{D}_\textit{female}$. For $\mathcal{D}_\textit{male}$ and $\mathcal{D}_\textit{female}$, we compute the respective unigram distributions $\mathcal{P}_{\textit{male}}$ and $\mathcal{P}_{\textit{female}}$. Next, for each token $t$, we compute the scores $\mathcal{P}_{\textit{male}}(t) - \mathcal{P}_{\textit{female}}(t)$, and $\mathcal{P}_{\textit{female}}(t) - \mathcal{P}_{\textit{male}}(t)$  and obtain the top tokens ranked by these scores (indicating increased usage in the respective sub-corpus). Table~\ref{tab:differential} indicates that male interviews are likelier to discuss technology, global politics, and sports than female interviews. In contrast, female interviews are likelier to discuss gender, family, and children as indicated by the presence of words \textit{girl}, \textit{woman}, \textit{gender}, and \textit{child}. We do not observe a single gendered word in the left column while we observe two gendered words in the right (e.g., \textit{woman} and \textit{girl}).

\begin{table}[htb]
\centering

\scriptsize
\begin{tabular}{| p{0.42\linewidth} | p{0.42\linewidth} |}
\hline
\textbf{More presence in $\mathcal{D}_\textit{male}$} & \textbf{More presence in $\mathcal{D}_\textit{female}$} \\
\hline
\textit{bengal, region, close, west, job, relative, happening, department, interest, industrial, accept, engineering, ukraine, agent, cricket, relation, option, subject, forest, iit}

&
\textit{woman, question, delhi, believe, capital, owner, important, something, good, deep, girl, education, place, gender, first, child, feel, science, health, doctor} \\
\hline
\end{tabular}
\vspace{.2cm}
\caption{Words with higher presence in $\mathcal{D}_\textit{male}$ (left) and $\mathcal{D}_\textit{female}$ (right).}
\label{tab:differential}
\end{table}



\subsubsection{Word Embedding Association Tests.}~\label{sec:weat}
Word Embedding Association Tests (\texttt{WEAT}) is a well-known framework to quantify gender bias~\cite{caliskan2017semantics} that has been widely used~\cite{lewis2020gender,DBLP:journals/patterns/KhadilkarKM22b}.  
Here, we are interested in answering the question -- \textit{does the candidate's gender matter in terms of discussion related to career and family?} We use \texttt{WEAT} score to quantify this association. Following prior literature~\cite{nosek2002harvesting}, we construct two target sets: \textit{Career} \{\texttt{executive, management, professional, corporation, salary, office, business, career}\} and \textit{Family} \{\texttt{home, parents, children, family, cousins, marriage, wedding, relatives}\}. We choose the following sets as the attributes representing gender qualifiers: \textit{Male} \{\texttt{male, man, he}\} and \textit{Female} \{\texttt{female, woman, she}\}. We train a well-known word embedding model (\texttt{FastText} \cite{bojanowski2017enriching}) on $\mathcal{D}$ to get the vectors corresponding to these words. Using these word vectors, we compute the WEAT score. Over five independent runs, we observe the \texttt{WEAT} score to be $0.29 \pm 0.07$. This positive score indicates a statistically significant association between males with career-oriented terms and females with family-oriented terms.

\subsubsection{Semantic Clustering of Questions by Topic.}\label{sec:semantic}

\begin{figure}[t]
  \centering
  \includegraphics[width=.7\columnwidth]{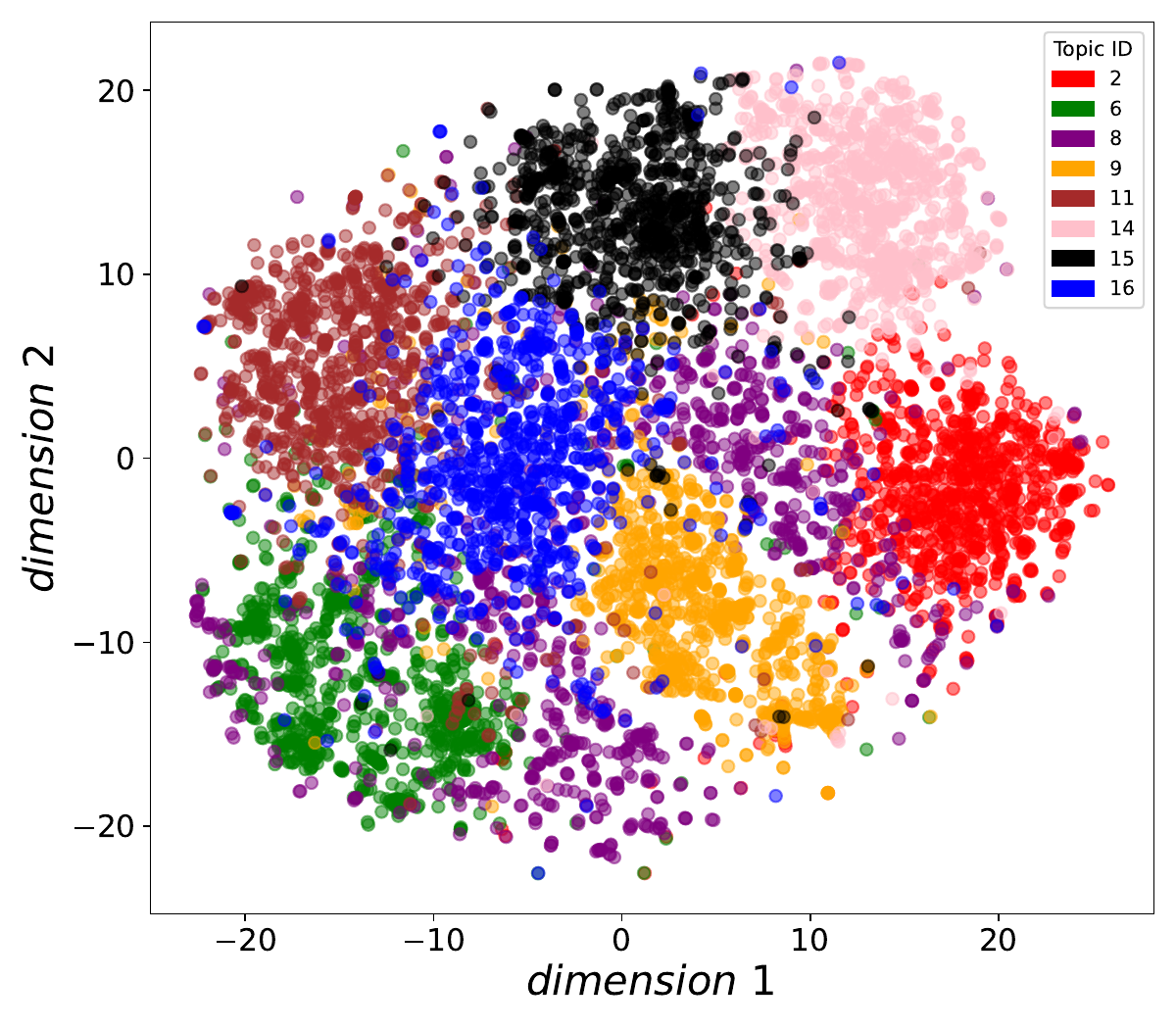} 
  \caption{t-SNE~\cite{van2008visualizing} visualization of top eight question topics. For better visualization, 1000 questions were randomly sampled from each cluster. Topic explanations -- 2: \textit{history and mythology}, 6: \textit{agriculture and environment}, 8: \textit{science}, 9: \textit{foreign policy}, 11: \textit{economics}, 14: \textit{gender related}, 15: \textit{law and order}, 16: \textit{engineering and technology}. Relevant keywords are listed in Table~\ref{table:clusters}.}
  \label{fig:tsne}
\end{figure}

To further study the differences in questioning patterns between male and female candidates, we cluster the questions into semantically similar topics. We compute the semantic embedding of the question texts in $\mathcal{Q}$ using a transformer-based embedding model, \texttt{all-MiniLM-L6-v2} \cite{reimers-2020-multilingual-sentence-bert}. We then run $K$-means \cite{wu2012advances}, an unsupervised clustering algorithm on these embeddings. The assumption here is that the clusters will have semantically similar questions. Initially, the number of clusters (topics) was set to 20. Among these, we are interested in the topics that exhibit a disparity in gender representation. To quantify this disparity, we use the imbalance ratio ($\mathcal{R}_\textit{imbalance}$) metric. For a topic $t$, the imbalance ratio is defined as 

$$\mathcal{R}_\textit{imbalance} = \frac{max\{f_\textit{male}^t, f_\textit{female}^t\}}{min\{f_\textit{male}^t, f_\textit{female}^t\}}$$

where $f_\textit{male}^t$ and $f_\textit{female}^t$denote the fraction of questions asked to male and female candidates, respectively, that belong to the topic $t$. In an ideal world where men and women candidates face similar questioning, the value of $\mathcal{R}_\textit{imbalance}$ should be $\sim1$ for any topic $t$. Conversely, a high $\mathcal{R}_\textit{imbalance}$ value indicates a significant skew in the distribution of questions toward one gender. Table \ref{table:clusters} presents the top eight topics displaying the greatest imbalance ratios. Figure \ref{fig:tsne} visualizes these topics. To better interpret the topics, we find the most prevalent phrases in each cluster and manually read a random sample of questions. Our analysis reveals that the questions related to \textit{gender equality} show the most skewed distribution, with female candidates nearly three times more likely to face such questions than their male counterparts. Among the other topics, questions related to \textit{agriculture and environment, engineering and technology, foreign policy, science,} and \textit{economics} were predominantly directed at male candidates, whereas \textit{history and mythology} and \textit{law and order} questions were more frequently posed to females. 

\begin{table*}[h!]
\centering
\scriptsize
\begin{tabular}{|c|c|c|c|c|p{7cm}|}
\hline
\rowcolor{gray!15}
\textbf{Topic ID} & 
$\mathbf{f_\textit{female}^t (\%)}$ & 
$\mathbf{f_\textit{male}^t (\%)}$ & 
$\mathcal{R}_\textit{imbalance}$ & 
\textbf{Topic Interpretation} & \textbf{Key Phrases} \\
\hline
\rowcolor{red!15} 14 & 4.19 & 1.43 & 2.94 & {gender equality} & \textit{woman, gender, female, empowerment, society, woman empowerment, sex ratio, gender equality, reservation woman, uttar pradesh, woman reservation, sexual harassment} \\
\hline
\rowcolor{blue!15} 6 & 3.68 & 4.49 & 1.33 & agriculture and environment & \textit{agriculture, farmer, forest, climate change, environment, pollution, sustainable development, global warming, renewable energy, development goal, power plant, green hydrogen, environmental issue, organic farming, rural area, food security, drinking water, green revolution, disaster management} \\
\hline
\rowcolor{blue!15} 16 & 4.78 & 5.96 & 1.25 & {engineering and technology} & \textit{big data, artificial intelligence, mechanical engineering, computer science, internet of things, machine learning, digital india, technology} \\
\hline
\rowcolor{red!15}2 & 7.90 & 6.43 & 1.23 & {history and mythology} & The key phrases in this cluster are not clear; however, a manual inspection reveals that questions are mostly related to history, mythology, and religious scriptures.\\
\hline
\rowcolor{blue!15}9 & 4.10 & 4.92 & 1.20 & {foreign policy} & \textit{international relation, foreign policy, prime minister, saudi arabia, united nation, european union, sri lanka, cold war, russia, ukraine, china, pakistan, afghanistan, taliban, world war, foreign trade, middle east, security} \\
\hline
\rowcolor{blue!15}8 & 5.84 & 6.87 & 1.18 & {science} & \textit{difference, virus, chemical, example, reason, basic difference} \\
\hline
\rowcolor{red!15}15 & 6.02 & 5.19 & 1.16 & {law and order} & \textit{law, supreme court, high court, fundamental right, constitutional amendment, district magistrate, information act, article, justice, police, rule, government} \\
\hline
\rowcolor{blue!15}11 & 5.90 & 6.53 & 1.11 & {economics} & \textit{economy, gdp, bank, budget, income tax, fiscal deficit, finance commission, growth rate, monetary policy, interest rate, stock market, demographic trend, fiscal policy, black money} \\
\hline
\end{tabular}%

\vspace{.2cm}
\caption{Descriptive analysis of question topics. Color coding: \colorbox{red!20}{Red} highlights topics with greater female representation, while \colorbox{blue!20}{Blue} signifies topics with greater male representation.}
\label{table:clusters}
\end{table*}

\begin{figure}[htb]
   \centering
   \includegraphics[width=.6\columnwidth]{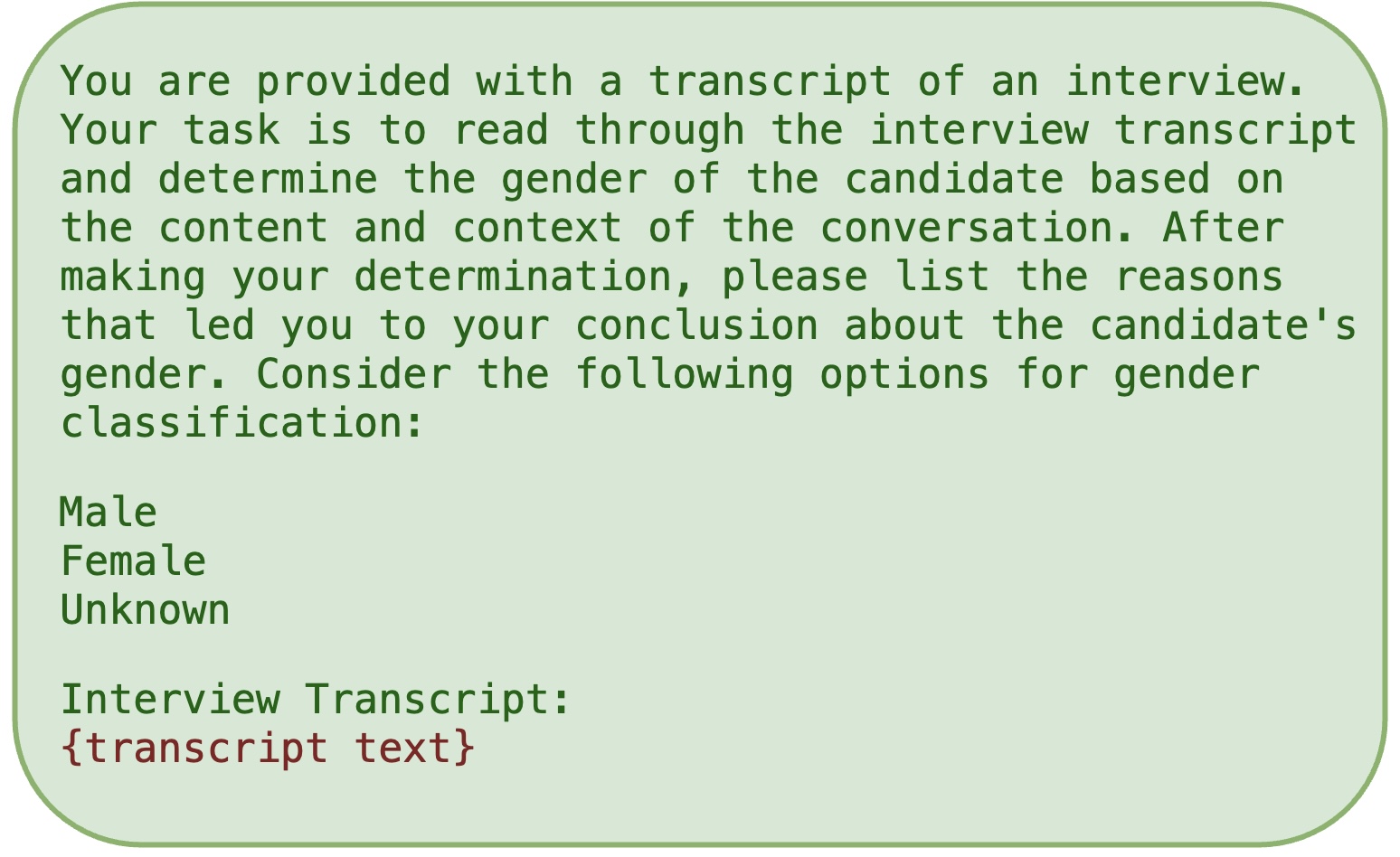}
   \caption{Prompt designed to infer gender using LLMs.}
   \label{fig:prompt}
\end{figure}

\subsection{Separability Tests}
The \textit{imbalance ratio} captures the differences in questioning patterns by analyzing the distribution of topics across male and female candidates. However, an important question still remains -- \textit{Does a systematic difference exist in the nature of questioning between the two groups regardless of the topic?} To this end, we conduct a set of separability tests. Following Dutta \textit{et al.} \cite{dutta2022murder}, we treat classification accuracy as a proxy for text separability. We construct the classification datasets by assigning labels to the questions from $\mathcal{Q}_\textit{female}$ (label \textbf{F}) and $\mathcal{Q}_\textit{male}$ (label \textbf{M}). Given a question, the classifier needs to predict whether it was asked to a female or male candidate. Intuitively, if there exist linguistic cues to differentiate between the questions, the task is learnable. However, if no such signals exist, the classifier will not perform better than chance. We describe the separability tests below.

Let $\mathcal{Q}_\textit{female}^t$ and $\mathcal{Q}_\textit{male}^t$ denote the questions asked to female and male candidates belonging to topic $t$. We sample an equal number of questions from $\mathcal{Q}_\textit{female}^t$ and $\mathcal{Q}_\textit{male}^t$, combine all topics, and split the data into train and test set in an 80:20 ratio. We fine-tune  \texttt{BERT} \cite{devlin2018bert}, a well-known pre-trained language model, for this classification task. As a control, we randomly divide $\mathcal{Q}_\textit{female}^t$ into two equal parts, combine all topics, and conduct the classification experiment. We repeat this process for $\mathcal{Q}_\textit{male}$. Table \ref{tab:sep1} presents results across test sets. We note that classification accuracy for $\mathcal{Q}_\textit{female}$ vs $\mathcal{Q}_\textit{male}$ is significantly higher than chance. Whereas the in-group classifiers perform no better than random guesses as expected. These results indicate that there exists a difference in the nature of questing between the male and female candidates.

\textbf{SI} contains additional experiments that show (1) linguistic separability of male-versus-female questions happens even if we control for topics (i.e., ensure each split has an equal number of questions from 
a given topic); and (2) linguistic separability of male-versus-female questions exists even within the same topic.


\begin{table}[htb]
\centering
\small
\setlength{\extrarowheight}{2pt}
\begin{tabular}{cc|c|c|}
\TBstrut
  & \multicolumn{1}{c}{} & \multicolumn{2}{c}{} \\
  & \multicolumn{1}{c}{} & \multicolumn{1}{c}{$\mathcal{Q}_\textit{female}$}  & \multicolumn{1}{c}{$\mathcal{Q}_\textit{male}$}  \\\cline{3-4}
  \TBstrut
            & $\mathcal{Q}_\textit{female}$ & \cellcolor{blue!25} 51.3 $\pm$ 0.9\%  & 58.0 $\pm$ 0.8\%
 \\ \cline{3-4}
 \TBstrut
  & $\mathcal{Q}_\textit{male}$  & 58.0 $\pm$ 0.8\% &\cellcolor{blue!25} 51.6 $\pm$ 0.8\%
 \\\cline{3-4}
\TBstrut
\end{tabular}
\caption{Separability test results between $\mathcal{Q}_\textit{female}$ and $\mathcal{Q}_\textit{male}$.}
\label{tab:sep1}
\end{table}

\vspace{-0.3cm}
\subsection{Bias in LLM Explanations for Gender Inference}
\label{llm_bias}
Here, we examine the explanations provided by the three LLMs (\texttt{Mistral-7B-Instruct}, \texttt{GPT-3.5-Turbo} and \texttt{Claude-3.5-Sonnet}) to assess whether the entrenched societal biases in language models significantly influence their gender inference. 

To infer gender using LLMs, we set up a zero-shot classification task with a prompt (Figure \ref{fig:prompt}) containing a detailed instruction followed by the interview transcript. We then extract the predicted gender and reasons from the response. It is worth noting that here, we do not include the candidate's name specifically; however, it may appear in the transcript if it is mentioned during the interview. Table \ref{tab:my_table} compares the performance of different LLMs on this inference tasks and shows that \texttt{Claude} demonstrates the highest performance in this task followed by \texttt{GPT} and \texttt{Mistral}.

We begin by extracting all rationales given by these models for each candidate, subsequently organizing these into two distinct datasets: $\mathcal{D}_\textit{reasons}^M$ for male predictions and $\mathcal{D}_\textit{reasons}^F$ for female predictions. We then analyze the unigram distribution within each dataset ($\mathcal{NF}_\textit{reasons}^M$ and $\mathcal{NF}_\textit{reasons}^F$) and perform a differential analysis similar to that described earlier. This analysis identifies words that are disproportionately frequent in one dataset compared to the other. Specifically, terms with high differential frequencies, $DF_{\mathcal{NF}_\textit{reasons}^M - \mathcal{NF}_\textit{reasons}^F}$, are indicative of a male prediction bias in LLM responses. Conversely, words with high scores in $DF_{\mathcal{NF}_\textit{reasons}^F - \mathcal{NF}_\textit{reasons}^M}$ suggest a female prediction bias. 

\begin{figure}[htb]
\centering
\begin{subfigure}{0.32\textwidth}
  \centering
  \includegraphics[frame, width=\linewidth]{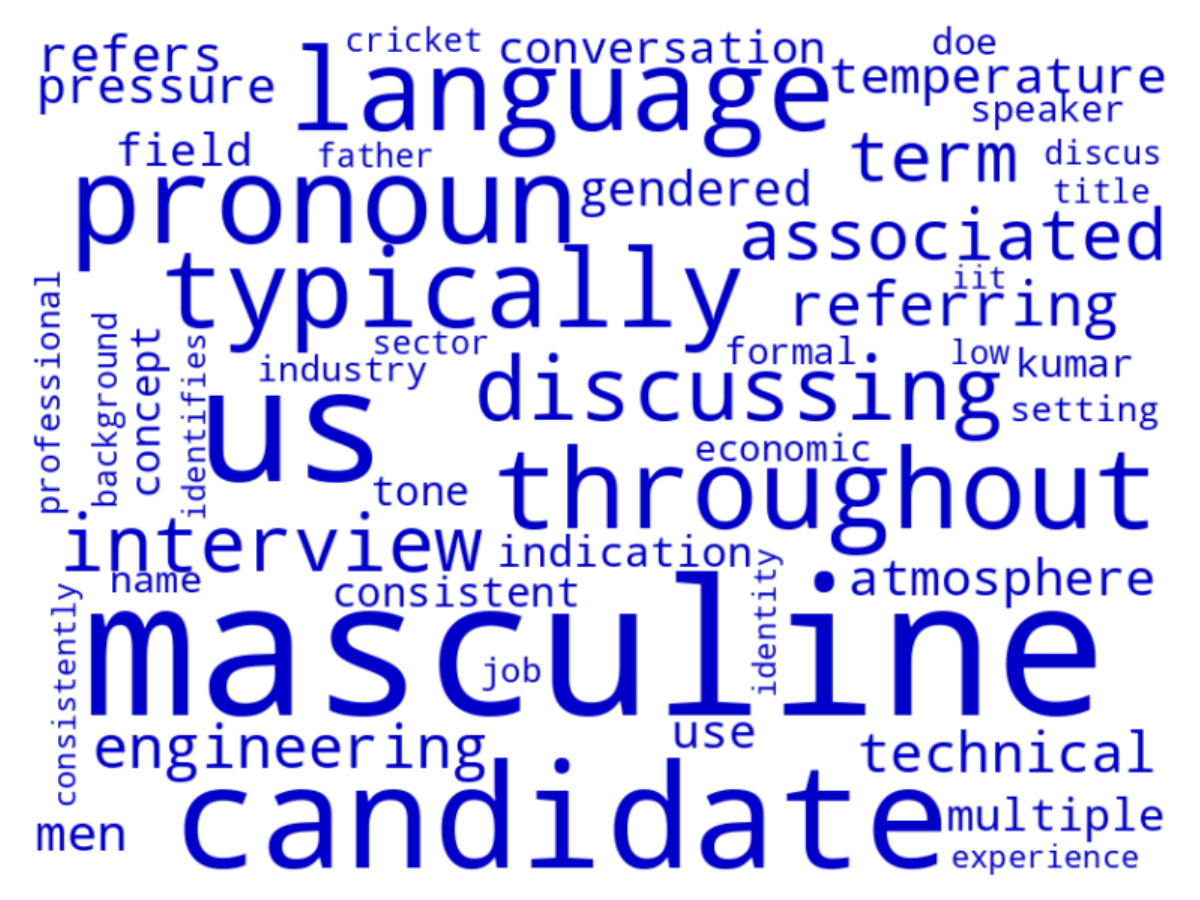}
  \caption{\texttt{Mistral-7B-Instruct}, male}
  \label{fig:sub1}
\end{subfigure}
\hfill
\begin{subfigure}{0.32\textwidth}
  \centering
  \includegraphics[frame, width=\linewidth]{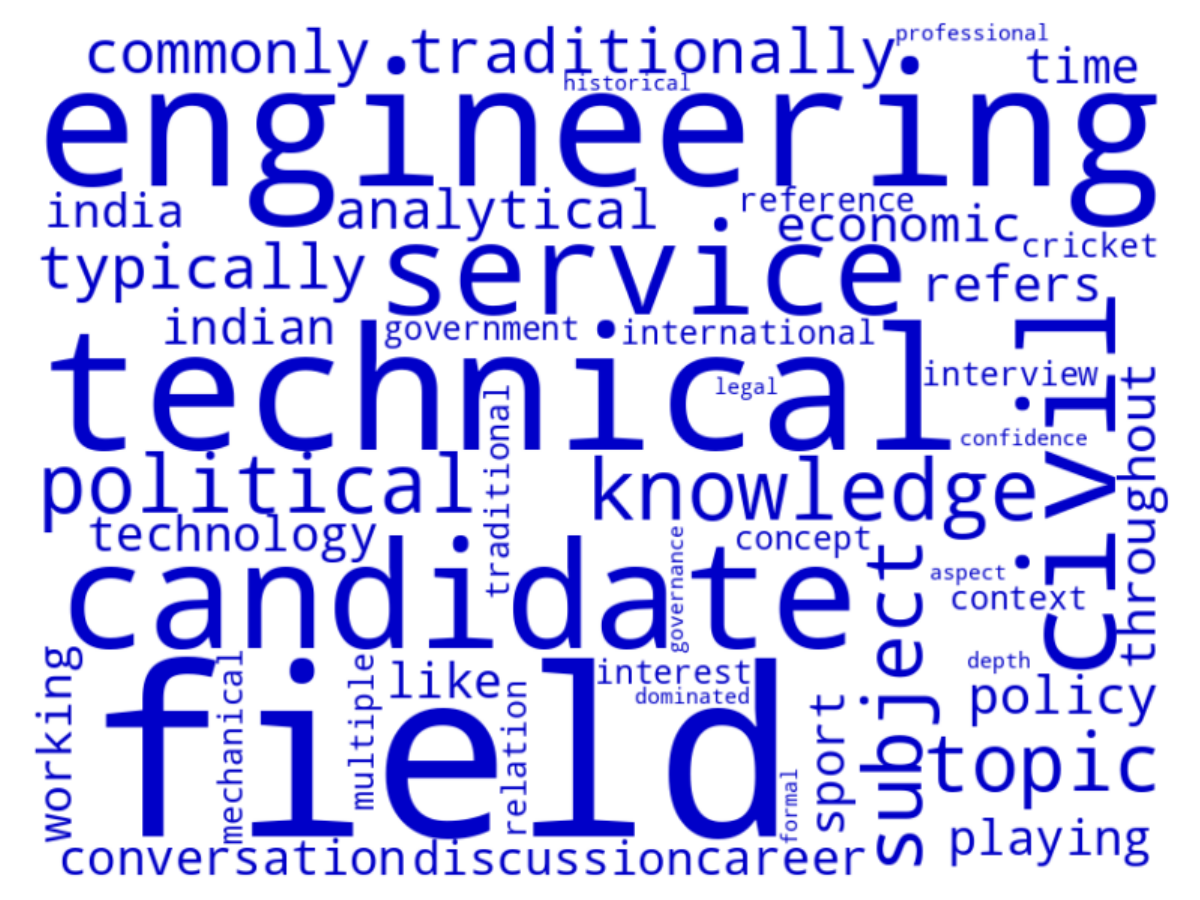}
  \caption{\texttt{GPT-3.5-Turbo}, male}
  \label{fig:sub2}
\end{subfigure}
\hfill
\begin{subfigure}{0.32\textwidth}
  \centering
  \includegraphics[frame, width=\linewidth]{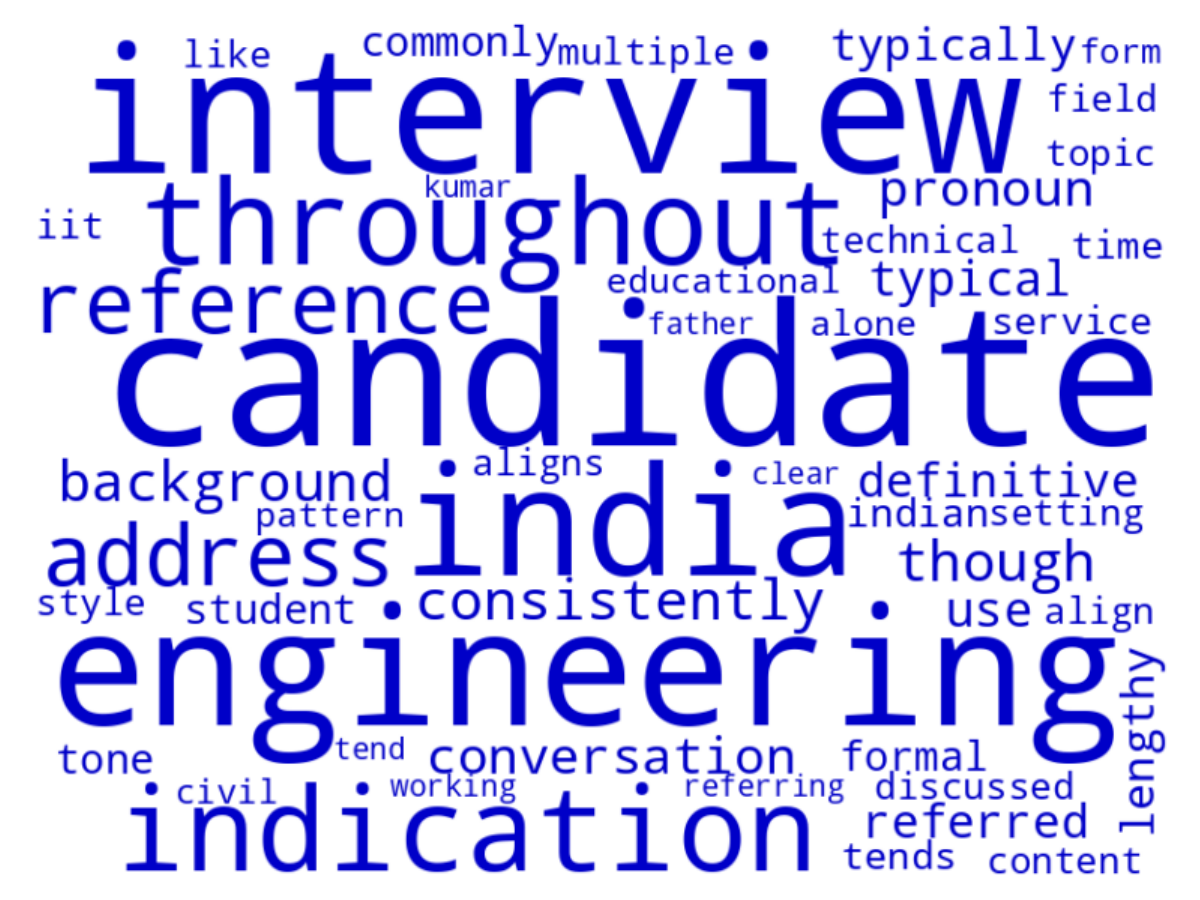}
  \caption{\texttt{Claude-3.5-Sonnet}, male}
  \label{fig:sub3}
\end{subfigure}

\begin{subfigure}{0.32\textwidth}
  \centering
  \includegraphics[frame, width=\linewidth]{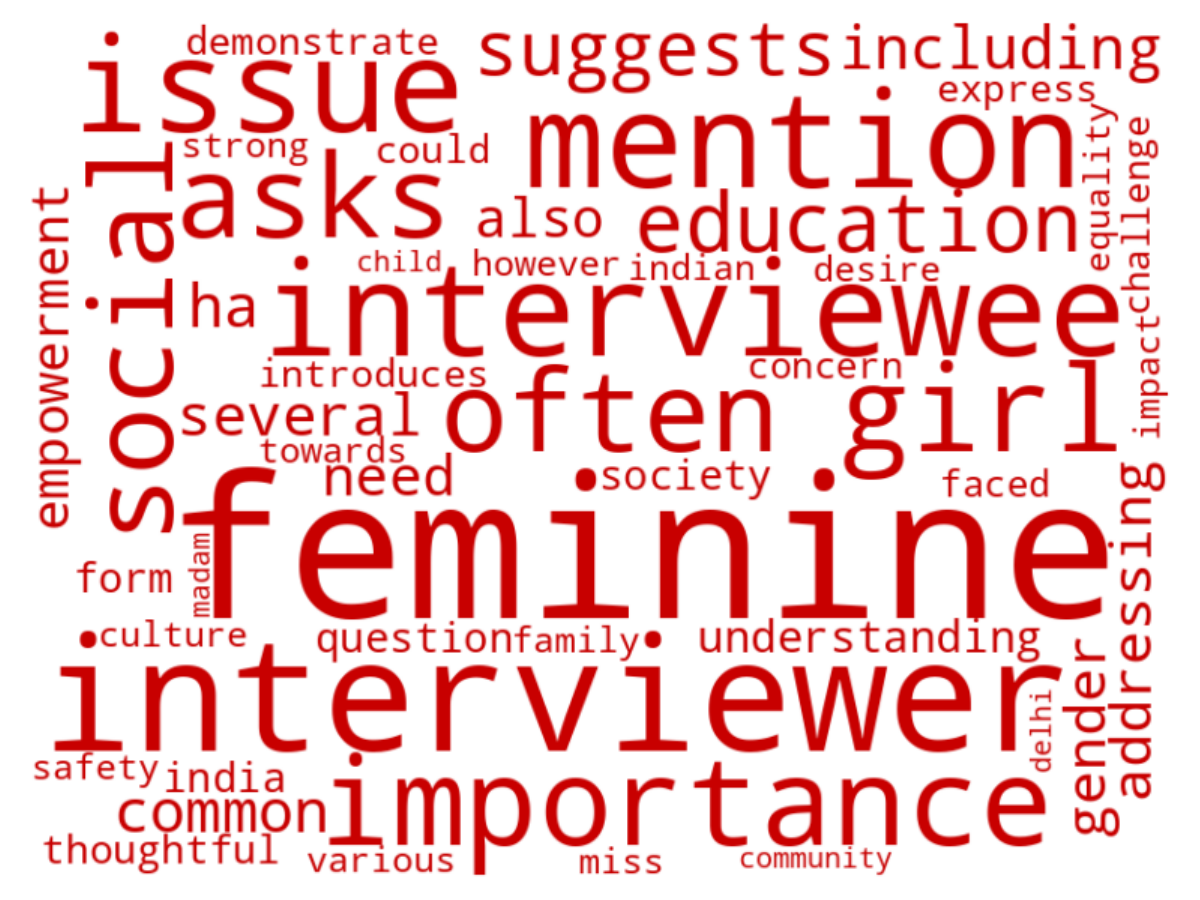}
  \caption{\texttt{Mistral-7B-Instruct}, female}
  \label{fig:sub4}
\end{subfigure}
\hfill
\begin{subfigure}{0.32\textwidth}
  \centering
  \includegraphics[frame, width=\linewidth]{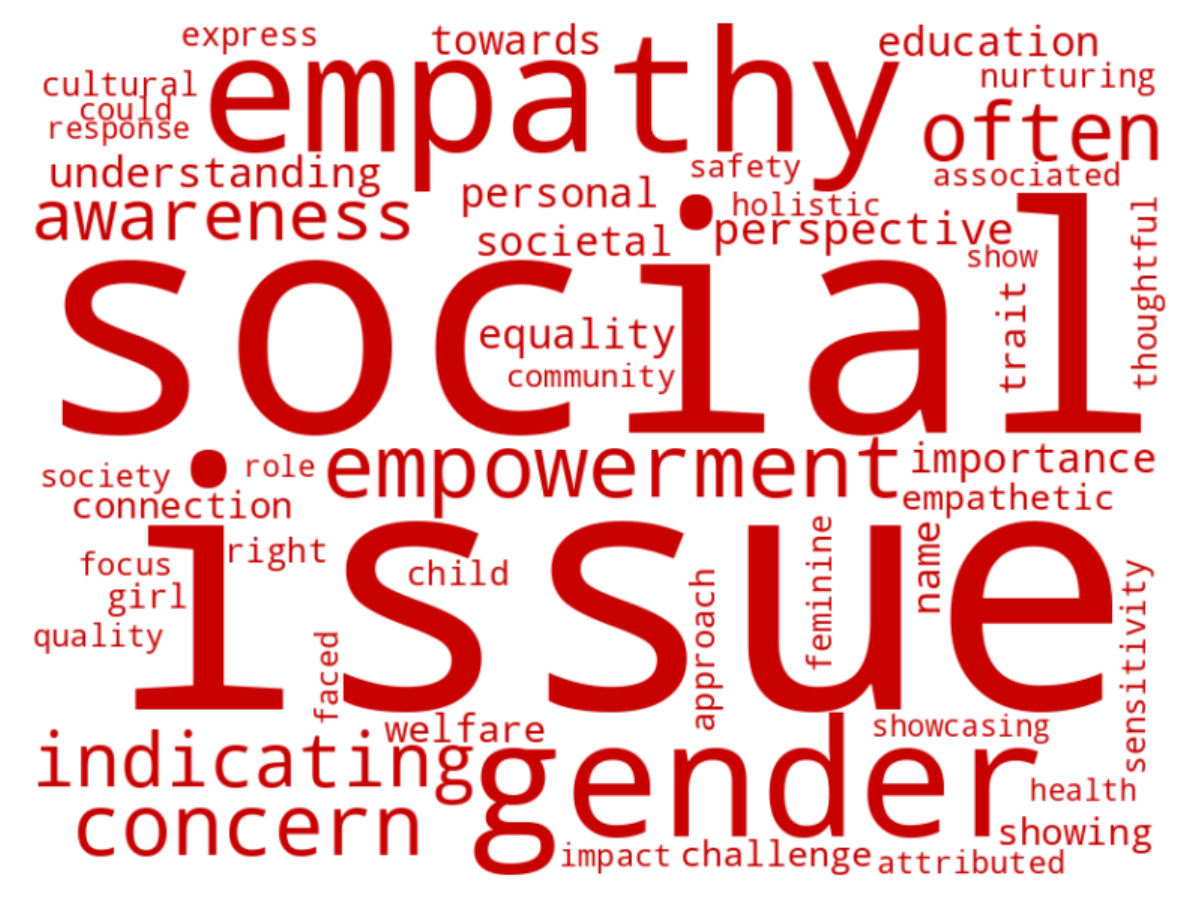}
  \caption{\texttt{GPT-3.5-Turbo}, female}
  \label{fig:sub5}
\end{subfigure}
\hfill
\begin{subfigure}{0.32\textwidth}
  \centering
  \includegraphics[frame, width=\linewidth]{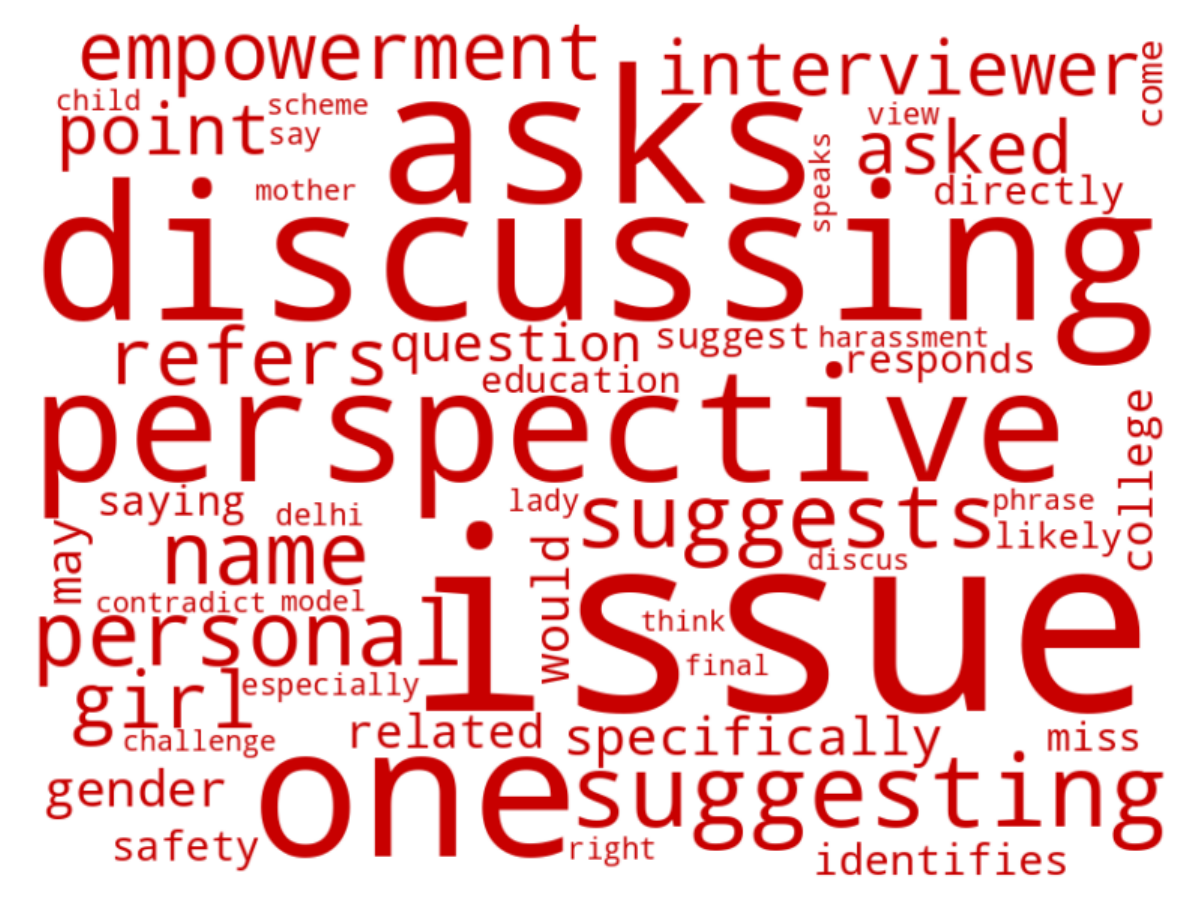}
  \caption{\texttt{Claude-3.5-Sonnet}, female}
  \label{fig:sub6}
\end{subfigure}

\caption{Wordclouds highlighting the top words found from the differential analysis of unigram distribution of LLM explanations. Top images illustrates words like \textit{engineering, technical, civil, knowledge} while the bottom images feature words like \textit{empathy, gender, social, issue, awareness} indicating the ingrained bias in the reasoning process of LLMs.}%
\label{fig:cloud}%
\end{figure}

We present six word clouds (Figure~\ref{fig:cloud}) that display the most significant words emerging from this differential analysis. The analysis reveals that all three models (Figures \ref{fig:cloud}a, \ref{fig:cloud}b, \ref{fig:cloud}c) are more likely to predict a candidate’s gender as male when encountering terms such as \textit{engineering, technical, civil} — words traditionally associated with male-dominated fields. In contrast, as shown in Figures \ref{fig:cloud}d, \ref{fig:cloud}e, \ref{fig:cloud}f), terms like \textit{gender, women empowerment, social issue} are predictive of a female gender identification by these models, underscoring potential biases in their training data that perhaps correlate these concepts with female gender. Interestingly, we note that if \texttt{GPT-3.5} finds qualities like \textit{empathy, awareness, and understanding} in the candidate's response, it predicts female. On the other hand, \texttt{Mistral} often determines a candidate's gender if it deems the language as masculine or feminine. Table \ref{tab:example_exp} lists a few examples showing the striking difference between the explanations provided by these LLMs for male and female candidates.

\begin{table}[h!]
\centering
\scriptsize
\begin{tabular}{|p{1.3cm}|l|p{4.8cm}|}
\hline
\rowcolor{blue!10}
\textbf{Predicted gender} & \textbf{LLM} & \textbf{LLM Explanation} \\
\hline
\rowcolor{gray!20}
Male & \texttt{GPT-3.5} & \textit{The candidate shows a {\textcolor{red}{strong knowledge of engineering concepts}}, which can be more commonly found in male candidates in technical fields.}\\
\hline
\rowcolor{white!20}
Female & \texttt{GPT-3.5} &\textit{The candidate's responses reflected \textcolor{red}{empathy, compassion}, and a focus on issues related to women empowerment, education, and societal challenges, which are often associated with female perspectives.} \\
\hline

\rowcolor{gray!20}
Male & \texttt{Mistral} & \textit{The candidate mentions his educational background, including his \textcolor{red}{M.Tech in transportation engineering} and his optional subject of \textcolor{red}{anthropology}, which are typically male-dominated fields.} \\
\hline

Female & \texttt{Mistral}  & \textit{... discusses issues related to the representation of tribal people and the \textcolor{red}{inclusion of women} in political and employment spheres, which are often topics of interest for female candidates.} \\
\hline

\rowcolor{gray!20}
Male & \texttt{Claude}  & \textit{The candidate discusses his \textcolor{red}{B.Tech degree in Mechanical Engineering}, a field that tends to have more male students.} \\
\hline

Female & \texttt{Claude} & \textit{The candidate is asked about procedures for \textcolor{red}{sexual harassment of women} in the workplace, which is a topic often directed at female candidates.} \\
\hline
\end{tabular}

\vspace{.2cm}
\caption{Examples demonstrating bias in the language models' rationale for gender predictions.}
\label{tab:example_exp}
\end{table}

\begin{table}[htb]
    \centering
 \small 
    \begin{tabular}{|c|c|}
        \hline
        \rowcolor{blue!15}
        \textbf{LLM} & \textbf{Cohen's $\kappa$} \\
        \hline
        \rowcolor{white!20}
        \texttt{Mistral-7B-Instruct}~\cite{jiang2023mistral} & 0.581 \\
        \rowcolor{gray!20}
        \texttt{GPT-3.5-Turbo} & 0.751 \\
        \rowcolor{white!20}
        \texttt{Claude-3.5-Sonnet} & 0.853 \\
        \hline
    \end{tabular}
    \vspace{.2cm}
    \caption{Gender inference evaluation. Cohen's $\kappa$ is computed with respect to $\mathcal{L}_\textit{comprehensive}$ inferred by humans.}
    \label{tab:my_table}
\end{table}

\section{Conclusion}

This paper conducts a comprehensive analysis of gender inequality through the lens of UPSC mock interview questions. UPSC is one of the most competitive exams in India and selected candidates form the administrative backbone of the world's largest democracy. Yet no prior literature (to our knowledge) has investigated gender inequality in mock interviews for one of the most high-profile government jobs in India. Our study is descriptive, not prescriptive. Our analyses reveal that while the interviewed female candidates are as strong as their male counterparts, their interview questions are strikingly different from the interview questions asked of the male candidates. We also observe that the interview panels are predominantly male. Finally, we present an intriguing finding that uncovers deep-seated gender bias in LLMs thoough the lens of a gender inference task.    

Our study raises the following thoughts.

\noindent\foo\textbf{\textit{Is women empowerment or gender equality solely a women's issue?}}  We observe female candidates are almost thrice as likely to be asked about gender empowerment as their male counterparts. In a world striving towards equality, do we expect equality only to be a topic for the disadvantaged group? Numbers indicate that women are catching up with men in UPSC representation. Numbers also indicate that they already match pound for pound when it comes to securing stellar ranks in the written exams. But women's improving performance notwithstanding, the questions they face in the interviews still reek of a patriarchal societal makeup. The skewed gender composition of the interview panels does not help either. If our findings from mock interviews also hold in actual UPSC interviews, we recommend restructuring the process toward gender neutrality. 

\noindent\foo\textbf{\textit{The lipstick gets darker and the pig more subtle, but the pig is still there:}} In their classic paper ``\textit{Lipstick on a Pig: Debiasing Methods Cover up Systematic Gender Biases
                  in Word Embeddings But do not Remove Them}'' Gonen and Goldberg~\cite{DBLP:conf/acl-wnlp/GonenG19} argued that debiasing methods merely provide a whitewash, while the deep-seated gender biases in the word embeddings remain. As we wade through the explanations presented by several cutting-edge LLMs on the gender inference task, we surprisingly notice that the lipstick has gotten darker, but the pig remains. Our findings thus indicate that we need rigorous bias audits of the new-age LLMs.   


\section{Limitations and Ethics Statement}

Transcription of the YouTube videos might not be the most accurate as models may introduce errors. We have used Whisper OpenAI in order to transcribe the videos. We have used proprietary LLMs such \texttt{GPT-3.5} and \texttt{Claude-Sonet-3.5}. Exact reproducibility of results might not be possible as the LLMs keep updating themselves. Our study investigates UPSC mock interview questions. While these mock interviews often invite experienced former IAS officers and noted academicians as panelists, it is not possible to estimate the fidelity of these mock interviews with the actual interviews. 

We collect public domain data using publicly available API. The interview candidates are highly visible public officials working at high-profile public-facing jobs. Instead of focusing on individual candidates, we conduct aggregate analyses. We thus see no major ethical concern. We rely on large language models for some of our analyses. Prior literature indicates possibilities of biases in these models which may percolate into downstream tasks. In fact, we report LLM biases in the explanations of the gender inference task which presents yet another data point in support of that concern. We verify our results through manual inspection whenever possible. Also, for some of our analyses (e.g., \texttt{WEAT} score), we train the word embeddings from scratch.  Our dataset also depends on the accuracy of the ASR system. Prior literature indicates such systems are not infallible~\cite{errattahi2018automatic}. Our manual inspection reveals that the quality of the transcriptions was high with occasional errors caused by the conversation switching to Hindi. We conduct additional preprocessing to account for that. 

Finally, any study on binary gender bias runs the risk of oversimplifying gender. We acknowledge that gender lies on a spectrum. We are also sensitive to previous studies that point out the potential harms of the erasure of gender and sexual minorities~\cite{ArjunErasurePaper}. It is possible that our gender inference has some noise. A recent global survey indicates that 3\% of the survey population identified as non-binary, non-conforming, gender-fluid, or transgender~\footnote{\url{https://www.statista.com/statistics/1269778/gender-identity-worldwide-country/}}. We induce a 3\% error in $\mathcal{L}_\textit{comprehensive}$ and observe that our qualitative claims remain unchanged (\textbf{SI} contains details). 

\bibliographystyle{plainnat}
\bibliography{article}

\appendix

\section{Word Embedding Association Test (\texttt{WEAT})}
Word Embedding Association Test (\texttt{WEAT}) \cite{caliskan2017semantics} score is a metric to detect if there exists a difference between two sets of target words in terms of their association with two sets of attribute words. To compute this metric, first, the words are converted to their vector representations (embeddings).
The cosine similarity of two words ($a$ and $b$) is denoted by $\cos(a, b)$. 

\begin{align*}
\text{WEAT}(\mathcal{X}, \mathcal{Y}, \mathcal{A}, \mathcal{B}) = & \ \text{mean}_{x \in \mathcal{X}} \sigma(x, \mathcal{A}, \mathcal{B}) \\
& - \text{mean}_{y \in \mathcal{Y}} \sigma(y, \mathcal{A}, \mathcal{B})
\end{align*}
where,
$$\sigma(w, \mathcal{A}, \mathcal{B}) = \text{mean}_{a \in \mathcal{A}} \cos(w, a) - \text{mean}_{b \in \mathcal{B}} \cos(w, b)$$

Intuitively, $\sigma(w, \mathcal{A}, \mathcal{B})$ quantifies the association of $w$ with the attribute sets, and the \texttt{WEAT} score measures the differential association of the two sets of target words with the attribute sets. A positive \texttt{WEAT} score suggests that the target words in set $\mathcal{X}$ have a stronger association with the attributes in set $\mathcal{A}$ than those in set $\mathcal{B}$, and conversely, the words in set $\mathcal{Y}$ show a stronger association with set $\mathcal{B}$ than with set $\mathcal{A}$.

\section{Log Odds Ratio Analysis}
We perform a log odds ratio analysis to find the words that are more likely to appear in female (male) interviews compared to their male (female) counterparts. The log odds of a word $w$ is defined as 
\small{
$$log~odds(w) = \frac{\text{normalized frequency of} ~w~ \text{in female interviews}}{\text{normalized frequency of} ~w~ \text{in male interviews}}$$}

\normalsize
A high positive value indicates that the word is more likely to appear in the interviews featuring a female candidate and a high negative value would indicate the opposite. We find the following words with high positive values of log odds -- \textbf{kathak} (dance form) (19.70), \textbf{rag} (musical structure) (3.81), \textbf{miranda} (house college) (3.16), nervous (2.28), \textbf{glass} (ceiling) (2.11), \textbf{sari} (2.10), \textbf{beauty} (1.83). On the other hand, these words show high negative scores -- \textbf{chess} (-19.10), \textbf{photography} (-18.71), \textbf{brexit} (-18.53), \textbf{automation} (-18.43) \textbf{football} (-2.80), \textbf{ncc} (National Cadet Corps) (-2.80), \textbf{camera} (-2.66), \textbf{alcohol} (-2.50).

The words with high association with women often indicate their hobbies and academic background. For instance, \textit{kathak} is one of the nine major Indian classical dance forms and \textit{rag} is a musical structure in Indian classical music. Since music is often mentioned as a hobby by the female candidates, we find this word's overpresence in female interviews. Miranda House College is a well-known gender-isolated college for women. We also observe words indicating female-traditional attire (sari). We were intrigued by the overpresence of the word \textit{glass}; manual inspection reveals that this word was used in the context of the phrase glass ceiling which further corroborates our earlier finding that gender inequality is more predominantly discussed with female candidates than with male candidates.  

With male candidates, we again observe that hobbies often dictated frequently used words. For instance, \textit{chess}, \textit{photography}, \textit{football}, and \textit{camera} were discussed in the context of hobbies. However, we do notice that words indicating world events (e.g., \textit{Brexit}) and technology (e.g., \textit{automation}) were present more frequently in male candidate interviews further substantiating our earlier finding that male candidates were more likely to be asked of questions about world politics and technology. 

Our findings point to a worrisome vicious cycle. On one hand, we do notice, that male candidates are asked about global politics more. But when a large language model uses an explanation during gender inference that a discussion heavy with world politics made it infer that the candidate is possibly male, may point to a problematic cycle where models learn from existing social biases and in turn, produce biased responses. 

\section{Background of Candidates}
Table \ref{tab:background} shows the academic backgrounds of UPSC candidates. Table~\ref{tab:academic-streams} contrasts the academic background distribution of a random sample of 200 candidates from $\mathcal{V}$ (obtained through manual inspection) with ground truth sourced from official UPSC data. Table~\ref{tab:academic-streams} establishes that $\mathcal{V}$ is a representative sample of successful UPSC candidates and is consistent with the academic distribution background of the recommended candidates.

\begin{table*}[htb]
\centering
\label{tab:heatmap}
\begin{tabular}{|>{\columncolor{gray!30}}c|c|c|c|c|}
\hline
\rowcolor{gray!30} 
 & \cellcolor{gray!30} Engineering & \cellcolor{gray!30} Humanities & \cellcolor{gray!30} Science & \cellcolor{gray!30} Medical Science \\
\hline
\cellcolor{gray!30} 2020 & \cellcolor{red!30} 64.9\% & \cellcolor{orange!30} 23.2\% & \cellcolor{yellow!30} 7.9\% & \cellcolor{green!30} 4\% \\
\hline
\cellcolor{gray!30} 2019 & \cellcolor{red!30} 63.1\% & \cellcolor{orange!30} 24.2\% & \cellcolor{yellow!30} 6.6\% & \cellcolor{green!30} 6.1\% \\
\hline

\cellcolor{gray!30} 2018 & \cellcolor{red!30} 62.7\% & \cellcolor{orange!30} 24.5\% & \cellcolor{yellow!30} 6.9\% & \cellcolor{green!30} 5.9\% \\
\hline

\cellcolor{gray!30} 2017 & \cellcolor{red!30} 66.2\% & \cellcolor{orange!30} 21.8\% & \cellcolor{yellow!30} 6.4\% & \cellcolor{green!30} 5.6\% \\
\hline

\cellcolor{gray!30} 2016 & \cellcolor{red!30} 59.3\% & \cellcolor{orange!30} 21.9\% & \cellcolor{yellow!30} 10.3\% & \cellcolor{green!30} 8.5\% \\
\hline

\end{tabular}

\caption{Academic Background of Recommended Candidates}
\label{tab:background}
\end{table*}

\begin{table}[htbp]
\small
    \centering
    \begin{tabular}{|l|p{2.2cm}| p{1.6cm}|}
        \hline
        \rowcolor{blue!10}
        Academic Stream & Distribution in $\mathcal{V}$ & Real World Distribution \\
        \hline
        Engineering & 63.05\% & 63.24\%\\
        \rowcolor{gray!15}
        Humanities & 25.10\% & 23.12\% \\
        Science & 4.45\%  & 7.62\% \\
        \rowcolor{gray!15}
        Medical Science & 7.40\% & 6.02\%\\
        \hline
    \end{tabular}
    \vspace{.2cm}
    \caption{Distribution of academic streams. Distribution in $\mathcal{V}$ is estimated by manually inspecting a random sample of 200 videos. Real world distribution is obtained from official UPSC data.}
    \label{tab:academic-streams}
\end{table}


\section{Question per Interview and Interview Time Duration}

We observe that male and female candidates receive comparable number of questions per interview (male candidates: 58.3 $\pm$ 22.1 questions; female candidates: 57.4 $\pm$ 20.8 questions) with male candidates receiving slightly more number of questions per interview. In a similar vein, we observe a male interview is marginally longer than a female interview (male interview: 30 minute 27 second $\pm$ 10 minute 7 second; female interview: 29 minute 25 second $\pm$ 9 minute 1 second).

\section{List of Channels and Channel Distribution}
\begin{figure}[htb]
   \textit{\centering}
   \includegraphics[width=.9\textwidth]{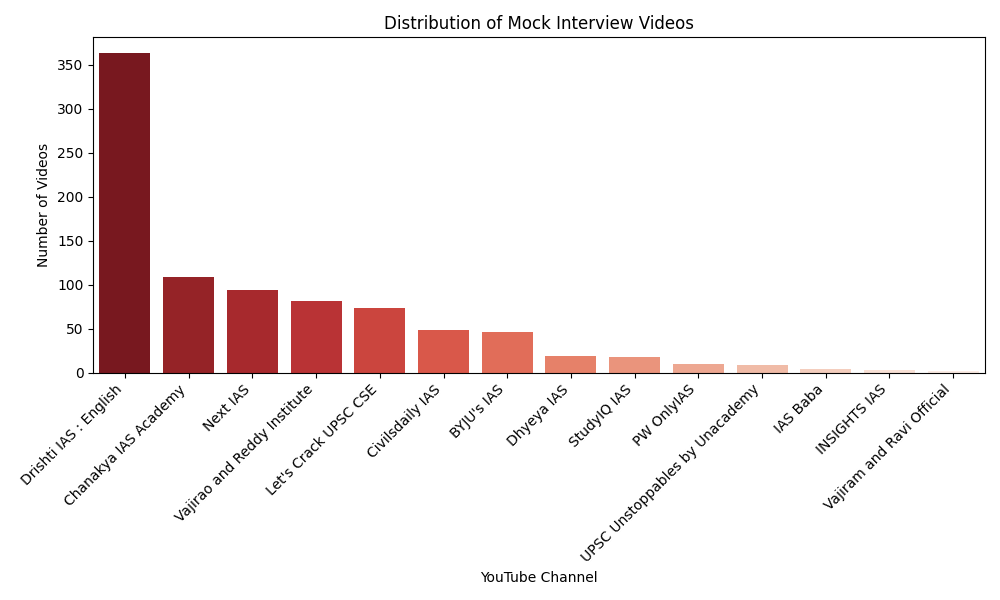}
   \caption{Distribution of mock interview videos across channels}
   \label{fig:channels}
\end{figure}

We use 14 well-known channels: \textit{Drishti IAS - English};
\textit{Chanakya IAS Academy};
\textit{Next IAS};
\textit{Vajirao and Reddy Institute};
\textit{Let's Crack UPSC CSE};
\textit{CivilsDaily IAS};
\textit{BYJU'S IAS};
\textit{Dhyeya IAS};
\textit{StudyIQ IAS};
\textit{PW OnlyIAS};
\textit{Unacademy};
\textit{IAS Baba};
\textit{INSIGHTS IAS}; and
\textit{Vajiram and Ravi Official}. Figure \ref{fig:channels} illustrates the distribution of channels in the dataset.

\section{FastText Training Parameters}
We use the following training parameters: 
\begin{itemize}
\item dimension = 100
\item epochs = 5
\item learning rate = 0.05
\item threads = 4.
\end{itemize}

\section{Topic Wise Separability Tests}
Table \ref{tab:sep2} lists the separability results for male-versus-female question classification within each topic.  

\begin{table}[htb]
\scriptsize
   \centering
   
   \begin{tabular}{|c|c|c|}
       \hline
       \rowcolor{blue!15}
       \textbf{Topic ID} & \textbf{Topic Interpretation} & \textbf{Classifier Accuracy} \\
       \hline
       \rowcolor{white!20}
       14 & \textit{gender equality} & 60.3\%\\
       \rowcolor{gray!20}
       6 & \textit{agriculture and environment} & 57.9\% \\
       \rowcolor{white!20}
       16 & \textit{engineering and technology} & 58.0\% \\
        \rowcolor{gray!20}
       2 & \textit{history and mythology} & 65.1\% \\
       \rowcolor{white!20}
       9 & \textit{foreign policy} & 55.3\% \\
        \rowcolor{gray!20}
       8 & \textit{science} & 62.5\% \\
       \rowcolor{white!20}
       15 & \textit{law and order} & 61.7\% \\
        \rowcolor{gray!20}
       11 & \textit{economics} & 55.9\% \\
    
       \hline
   \end{tabular}

   \vspace{.2cm}
   \caption{Separability test results within topics. This result demonstrates that even when we consider a specific topic, questions asked of male candidates are linguistically different from questions asked of female candidates.}
    
   \label{tab:sep2}
\end{table}

\section{Cluster Analysis with Noise}
We acknowledge that our human gender inference could have errors. Following a recent global survey that indicates that nearly 3\% of the survey population identified as non-binary, non-conforming, gender-fluid, or transgender~\footnote{\url{https://www.statista.com/statistics/1269778/gender-identity-worldwide-country/}}, we induce a 3\% error in $\mathcal{L}_\textit{comprehensive}$ and observe that our qualitative claims remain unchanged. 
Table \ref{table:clusters_noise} presents the analysis of different question topics after introducing 3\% noise in $\mathcal{L}_\textit{comprehensive}$.

\begin{table}[]
\centering
\scriptsize
\begin{tabular}{|c|c|c|c|c|p{8cm}|}
\hline
\rowcolor{gray!15}
\textbf{Topic ID} & 
$\mathbf{f_\textit{female}^t (\%)}$
& $\mathbf{f_\textit{male}^t (\%)}$
& $\mathcal{R}_\textit{imbalance}$ 
& \textbf{Topic Interpretation} 
& \textbf{Key Phrases} \\
\hline
\rowcolor{red!15} 14 & 4.08 & 1.44 & 2.83 & {gender equality} & \textit{woman, gender, female, empowerment, society, woman empowerment, sex ratio, gender equality, reservation woman, uttar pradesh, woman reservation, sexual harassment} \\
\hline
\rowcolor{blue!15} 6 & 3.77 & 4.88 & 1.29 & agriculture and environment & \textit{agriculture, farmer, forest, climate change, environment, pollution, sustainable development, global warming, renewable energy, development goal, power plant, green hydrogen, environmental issue, organic farming, rural area, food security, drinking water, green revolution, disaster management} \\
\hline
\rowcolor{blue!15} 16 & 4.78 & 5.98 & 1.25 & {engineering and technology} & \textit{big data, artificial intelligence, mechanical engineering, computer science, internet of things, machine learning, digital india, technology} \\
\hline
\rowcolor{red!15}2 & 7.89 & 6.41 & 1.23 & {history and mythology} & The key phrases in this cluster are not clear; however, a manual inspection reveals that questions are mostly related to history, mythology, and religious scriptures.\\
\hline
\rowcolor{blue!15}9 & 4.06 & 4.96 & 1.22 & {foreign policy} & \textit{international relation, foreign policy, prime minister, saudi arabia, united nation, european union, sri lanka, cold war, russia, ukraine, china, pakistan, afghanistan, taliban, world war, foreign trade, middle east, security} \\
\hline
\rowcolor{blue!15}8 & 5.88 & 6.87 & 1.17 & {science} & \textit{difference, virus, chemical, example, reason, basic difference} \\
\hline
\rowcolor{red!15}15 & 6.09 & 5.13 & 1.19 & {law and order} & \textit{law, supreme court, high court, fundamental right, constitutional amendment, district magistrate, information act, article, justice, police, rule, government} \\
\hline
\rowcolor{blue!15}11 & 5.84 & 6.58 & 1.13 & {economics} & \textit{economy, gdp, bank, budget, income tax, fiscal deficit, finance commission, growth rate, monetary policy, interest rate, stock market, demographic trend, fiscal policy, black money} \\
\hline
\end{tabular}%

\caption{Descriptive analysis of question topics (with added noise). Color coding: \colorbox{red!20}{Red} highlights topics with greater female representation, while \colorbox{blue!20}{Blue} signifies topics with greater male representation.}
\label{table:clusters_noise}
\end{table}

\end{document}